\documentclass[conference]{IEEEtran}
\usepackage{times}
\IEEEoverridecommandlockouts% This command is only needed if you want to use the \thanks command

\usepackage[numbers]{natbib} % template
\usepackage{multicol} % template
\usepackage[bookmarks=true,pdfencoding=auto,breaklinks,colorlinks,hypertexnames=false]{hyperref} % template
\usepackage{graphicx}
\usepackage{amsmath}
\usepackage{amssymb}
\usepackage{booktabs}
\usepackage{xcolor}
\usepackage[flushleft]{threeparttable}
\usepackage{multirow, bigdelim}
\usepackage{url}
\usepackage[export]{adjustbox}

\usepackage{siunitx}
\usepackage{flushend}
\usepackage{colortbl}
\usepackage{pifont}
\usepackage{cuted}
\usepackage{tcolorbox}

\usepackage{caption}
\usepackage{subcaption}
\usepackage{pgfplots}
\pgfplotsset{width=7cm, compat=1.10}
\usepgfplotslibrary{fillbetween}

\definecolor{dark-green}{RGB}{12,80,12}

\renewcommand{\eqref}[1]{Eq.~(\ref{#1})}

\usepackage{glossaries}
\newacronym{ml}{ML}{Machine Learning}
\newacronym{dl}{DL}{Deep Learning}
 \newacronym{cnn}{CNN}{Convolutional Neural Network}
 \newacronym{lstm}{LSTM}{Long Short-Term Memory}
 \newacronym{svm}{SVM}{support Vector Machine}
 \newacronym{rcnn}{RCNN}{Region-Based Convolutional Neural Network}
 \newacronym{fpn}{FPN}{Feature Pyramid Networks}
 \newacronym{yolo}{YOLO}{You Only Look Once}
 \newacronym{ssd}{SSD}{Single-Shot Multibox Detector}
 \newacronym{detr}{DETR}{Detection Transformer}
 \newacronym{nlp}{NLP}{Natural Language Processing}
 \newacronym{nms}{NMS}{Non-Maximum Suppression}
 \newacronym{lgt}{LGT}{Lane Graph Transformer}
 \newacronym{ffn}{FFN}{Feed Forward Network}
 \newacronym{fcn}{FCN}{Fully Convolutional Network}
 \newacronym{pspnet}{PSPNet}{Pyramid Scene Parsing Network}
 \newacronym{deconvnet}{DeConvNet}{Deconvolution Network}
 \newacronym{resnet}{ResNet}{Residual Network}
 \newacronym{aspp}{ASPP}{Atrous Spatial Pyramid Pooling}
 \newacronym{rnn}{RNN}{Recurrent Neural Network}
 \newacronym{hd}{HD}{High Definition}
 \newacronym{lidar}{LiDAR}{Light Detection And Ranging}
 \newacronym{dag}{DAG}{directed acyclic graph}
 \newacronym{pv}{PV}{perspective view}
 \newacronym{bev}{BEV}{bird's eye view}
 \newacronym{gnn}{GNN}{Graph Neural Network}
 \newacronym{lbfgs}{L-BFGS}{Limited-memory Broyden–Fletcher–Goldfarb–Shanno}
 \newacronym{adas}{ADAS}{Advanced Driver Assistance Systems}
 \newacronym{iou}{IoU}{Intersection over Union}
 \newacronym{apls}{APLS}{Average Path Length Similarity}
 \newacronym{sda}{SDA}{Split Detection Accuracy}
 \newacronym{dfs}{DFS}{Depth-First-Search}
 \newacronym{vit}{ViT}{Vision Transformer}
 \newacronym{mlp}{MLP}{Multi Layer Perceptron}
 \newacronym{bce}{BCE}{Binary Cross Entropy}
 \newacronym{mse}{MSE}{Mean Squared Error}
 \newacronym{ddp}{DDP}{Distributed Data Parallel}
\glsdisablehyper

\begin{document}

\title{Learning Lane Graphs from Aerial Imagery \\ Using Transformers}

% You will get a Paper-ID when submitting a pdf file to the conference system
% \author{Author Names Omitted for Anonymous Review. Paper-ID [add your ID here]}

\author{\authorblockN{
Martin Büchner\textsuperscript{*},\quad
Simon Dorer\textsuperscript{*},\quad
Abhinav Valada}
\vspace{0.2cm}
\authorblockA{University of Freiburg}
\thanks{$^{*}$ Equal contribution.}%
\thanks{This work was funded by the German Research Foundation (DFG) Emmy Noether Program grant number 468878300 and an academic grant from NVIDIA.}%
}

% avoiding spaces at the end of the author lines is not a problem with
% conference papers because we don't use \thanks or \IEEEmembership

%Homer Simpson\authorrefmark{2},
%Montgomery Scott\authorrefmark{3} and
%Eldon Tyrell\authorrefmark{4}}
%Telephone: (800) 555--1212, Fax: (888) 555--1212}
%\authorblockA{\authorrefmark{4}Tyrell Inc., 123 Replicant Street, Los Angeles, California 90210--4321}}

\maketitle

\begin{abstract}
    The robust and safe operation of automated vehicles underscores the critical need for detailed and accurate topological maps. At the heart of this requirement is the construction of lane graphs, which provide essential information on lane connectivity, vital for navigating complex urban environments autonomously. While transformer-based models have been effective in creating map topologies from vehicle-mounted sensor data, their potential for generating such graphs from aerial imagery remains untapped. This work introduces a novel approach to generating successor lane graphs from aerial imagery, utilizing the advanced capabilities of transformer models. We frame successor lane graphs as a collection of maximal length paths and predict them using a Detection Transformer (DETR) architecture. We demonstrate the efficacy of our method through extensive experiments on the diverse and large-scale UrbanLaneGraph dataset, illustrating its accuracy in generating successor lane graphs and highlighting its potential for enhancing autonomous vehicle navigation in complex environments.

\end{abstract}

\IEEEpeerreviewmaketitle

\section{Introduction}
Modern self-driving vehicles rely on accurate topological representations of their immediate surroundings to plan and navigate safely. While onboard sensors such as cameras or LiDARs provide vital observations used in localization~\cite{greve2023collaborative, arce2023padloc}, scene understanding~\cite{buchner20223d, mohan2022perceiving, gosala2023skyeye} and downstream decision making~\cite{trumpp2023efficient}, they come with significant challenges such as limited availability and occlusions. Nonetheless, robust planning and control require high-fidelity topologies such as HD map data to act in a timely and safe manner. Thus far, both LiDARs and cameras have been successfully used for predicting the road topology~\cite{maptr,maptrv2,hdmapnet} in the vicinity of the ego vehicle. However, the input sensor data is starkly affected by occlusions leading to frequent hallucinations, which is common for supervised learning techniques.

\begin{figure}
    \centering
    \includegraphics[width=1.0\linewidth]{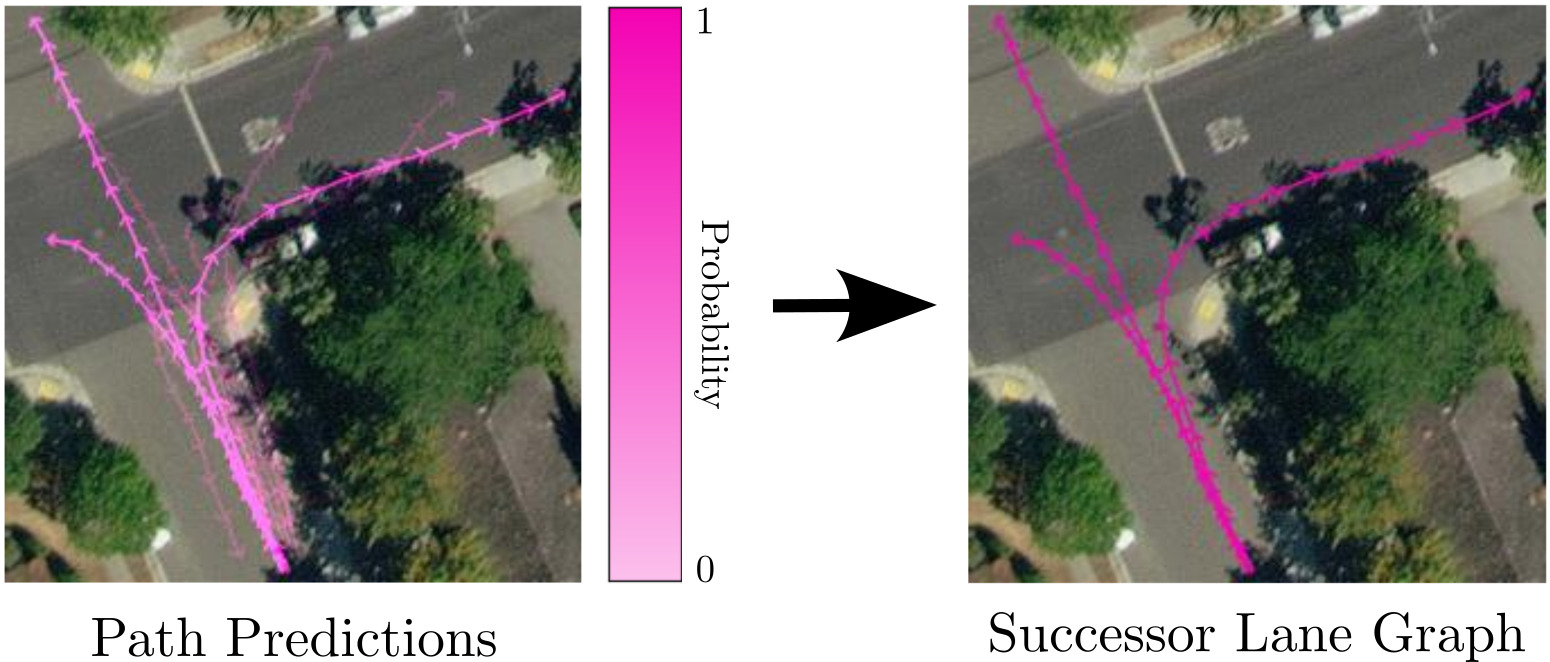}
    \caption{We present aerial lane graph transformers (ALGT) for learning feasible high-fidelity traversals of successor lane graphs. The left image shows raw predicted traversals while their opacity represents the predicted probability score. The right image shows the thresholded and aggregated traversals forming a successor lane graph.}
    \label{fig:teaser}
    \vspace{-0.5em}
\end{figure}

Lane graphs are a fundamental element of topological maps as they provide lane-level connectivity information of the road network suitable for planning, trajectory prediction, and control. However, predicting a sparse data structure such as a lane graph is usually hard due to its connectivity~\cite{lanegap, lane-graph-net}. In order to resolve occlusions in the surround view, several methods have been recently proposed that leverage aerial and bird's-eye-view (BEV) data for improving the lane graph prediction accuracy~\cite{street-map-extraction, lane-gnn, lane-graph-net}. While these aerial methods provide useful representations, they either lack capabilities in out-of-distribution scenarios~\cite{street-map-extraction}, suffer from inaccurate connectivity~\cite{lane-graph-net}, or cannot provide the required level of fidelity of, e.g., the node positions~\cite{lane-gnn}.

In this work, we tackle the problem of predicting successor lane graphs, as introduced in the \textit{UrbanLaneGraph} dataset~\cite{lane-gnn}. Successor lane graphs provide actionable representations given an initial starting pose of the ego-vehicle useful for short-term planning as depicted in Fig.~\ref{fig:teaser}. For this task, we follow Liao et al.~\cite{lanegap} in decomposing the graph-level task into multiple path-level predictions. Leveraging recent developments in set-based prediction (DETR)~\cite{detr,dab-detr, deformable-detr}, we introduce the Aerial Lane Graph Transformer (ALGT) model. Our proposed network takes aerial images including context as input to predict successor lane graphs within a region of interest at the center of the input image as shown in Fig.~\ref{fig:dataset-viz}. The successor lane graphs themselves are predicted as multiple feasible traversals of a directed-acyclic graph, which are later fused to form final predictions. As part of this work, we demonstrate the benefit of using polyline path representations over Bézier parametrizations for improved graph prediction. In addition, we present insights into the capabilities of various image backbones used for encoding aerial imagery.

To summarize, our main contributions are:
\begin{itemize}
    \item We present the ALGT framework for predicting successor lane graphs from aerial imagery using transformers.
    \item We demonstrate the benefit of using polyline representations over Bézier parametrization and ablate over various network architectures.
\end{itemize}

\section{Technical Approach}

Our approach aims to predict successor lane graphs $\hat{\mathcal{G}}$ in terms of path proposals $\hat{Y}$ given aerial image crops $\mathbf{I}$ in a supervised learning manner. In the following, we outline two lane graph representations (Sec.~\ref{subsec:lane-graph-repr}), introduce the ALGT model in Sec.~\ref{subsec:aerial-lane-grap-transformer}, detail the training procedure in Sec.~\ref{subsec:training}, and describe the aggregation of proposal paths to form a single successor graph $\hat{\mathcal{G}}$ in Sec.~\ref{subsec:path-aggregation}.

\subsection{Lane Graph Representation}
\label{subsec:lane-graph-repr}
\begin{figure}[h]
    \centering
    \includegraphics[width=0.22\textwidth]{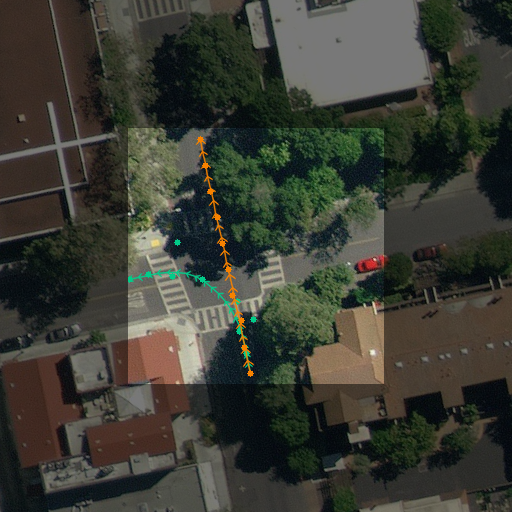}
    \includegraphics[width=0.22\textwidth]{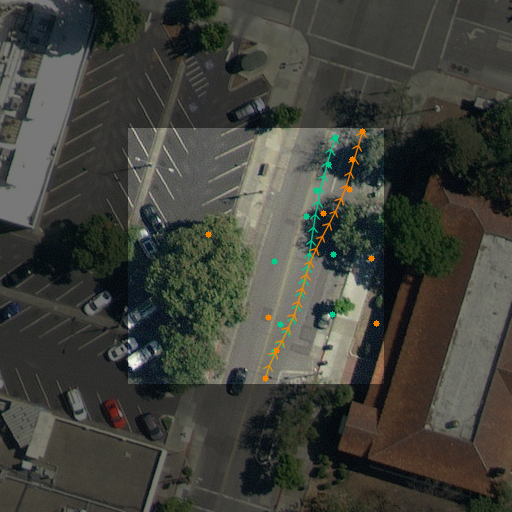}\\[0.15em]
    \includegraphics[width=0.22\textwidth]{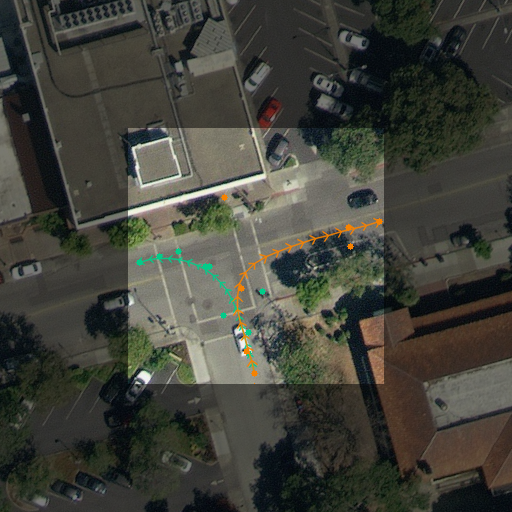}
    \includegraphics[width=0.22\textwidth]{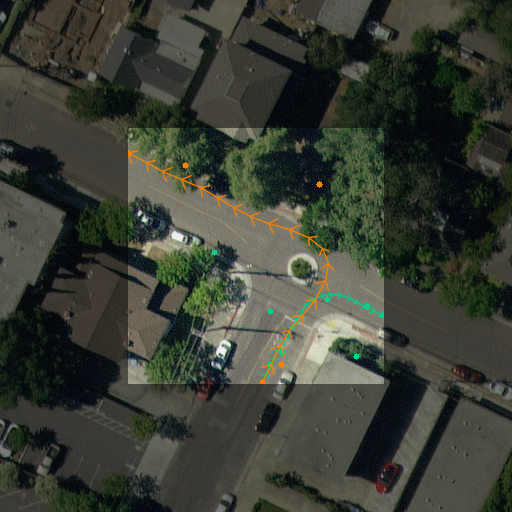}   
    \caption{Visualization of decomposed successor lane graphs of the UrbanLaneGraph dataset \cite{lane-gnn}. We choose Bézier curves of degree 10 and polylines from 20 sample points on this curve. All paths along with their Bézier control points are depicted in varying colors. The context part of the samples is darkened.}
    \label{fig:dataset-viz}
\end{figure}

We adopt the path-level representation strategy as proposed by \textit{MapTR}~\cite{maptr} and decompose $\mathcal{G}$ into a set of maximal length paths. These paths represent a collection of traversals from the initial ego pose $v_0$ to arbitrary terminal nodes $v_{end}$ of $\mathcal{G}$. This path-level decomposition is necessary for predicting set-wise outputs later using the transformer architecture. We can represent these traversal paths either as polylines or as Bézier curves:

\begin{itemize}
    \item A polyline \(\mathcal{P}_n\) consists of $n$ straight line segments linked sequentially to create a piece-wise linear path. It is defined by a series of points \(\boldsymbol p_0, \boldsymbol p_1, \ldots, \boldsymbol p_k\) located in a two-dimensional space. Each adjacent pair of points \((\boldsymbol p_{i}, \boldsymbol p_{i+1})\) is connected by a straight line.
    \item A Bézier curve $\mathcal{B}_n$ in \(n\) dimensions is specified by \(n + 1\) control points \(\boldsymbol b_0, \boldsymbol b_1, \ldots, \boldsymbol b_{n}\), with \(\boldsymbol b_0\) marking the start and \(\boldsymbol b_{n}\) the end of the curve. The intermediate control points influence the curve's shape through Bernstein polynomials of the form \(B_{i, n} = \binom{n}{i}t^i (1 - t)^{n - i}\). A Bézier curve is formulated as
\begin{equation}
    \mathcal{B}_{n}(t) = \sum_{i = 0}^{n} \boldsymbol b_i B_{i, n}(t) \quad \text{for} \quad 0 \le t \le 1.
\end{equation}
\end{itemize}
We visualize these decomposed path representations including the Bézier control points in Fig.~\ref{fig:dataset-viz}.

\begin{figure*}
    \centering
    \includegraphics[width=1.0\textwidth]{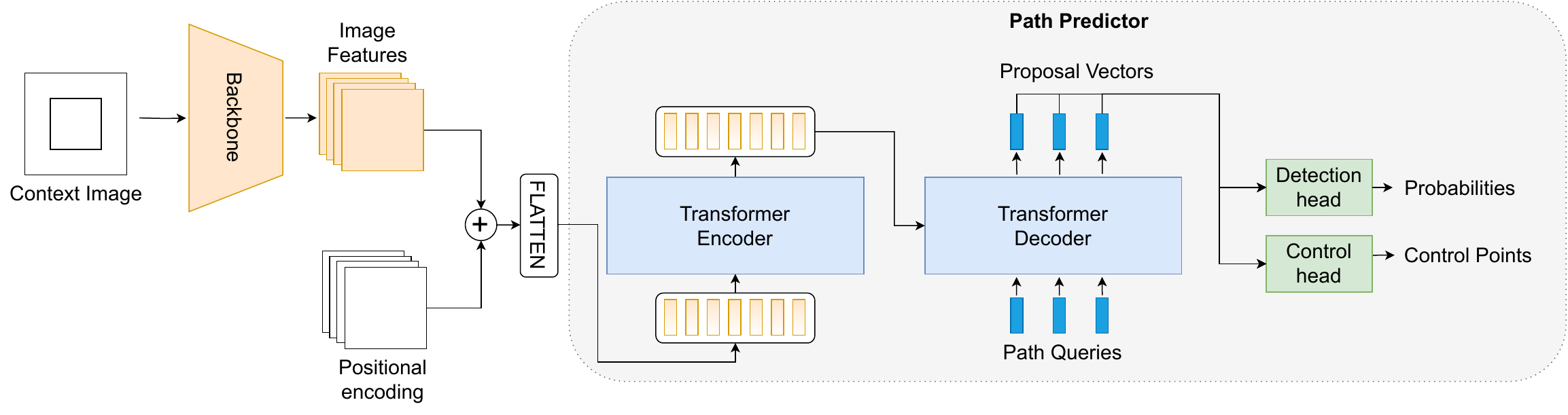}
    \caption{Overview of our ALGT model for successor lane graph prediction. An image backbone extracts relevant features from the context image that are passed through a transformer-based path predictor to produce successor lane graph proposals.}
    \label{fig:approach}
\end{figure*}

\subsection{Aerial Lane Graph Transformer}
\label{subsec:aerial-lane-grap-transformer}
The input to the proposed transformer model is an RGB image crop $\boldsymbol{I}$, from which the central region represents the actual target image $\boldsymbol I_{roi}$, while the outer margin serves as contextual information. We aim to predict a successor lane graph $\hat{\mathcal{G}}$ within the image domain $\boldsymbol I_{roi}$, where the initial node $v_0$ is positioned at the bottom center of $\boldsymbol I_{roi}$. This is also visualized in Fig.~\ref{fig:dataset-viz} where the dark region represents the context and the inner region represents $\boldsymbol I_{roi}$. Our ALGT model is structured around three primary components. Initially, an image backbone extracts relevant features from the input image \(\boldsymbol{I}\) (Sec.~\ref{subsec:backbone}). Following this, a transformer-based path predictor derives a set of path proposals \(\hat{Y}\) (Sec.~\ref{subsec:path-predictor}). These proposals are later aggregated. An overview of our model's architecture is illustrated in Fig.~\ref{fig:approach}.

\subsubsection{Aerial Image Backbone}
\label{subsec:backbone}
To extract features from the context RGB image \(\boldsymbol{I}\), we leverage an ego lane regression network, similar to LaneGNN~\cite{lane-gnn}. The network is built upon two stacked PSPNet~\cite{pspnet} instances, initially generating a context lane regression mask. In combination with the original input image, the context regression mask is then used to predict an ego lane regression mask. We disregard the PSPNet's upsampling layers and prediction heads and use it as our image backbone, transforming \(\boldsymbol{I}\) into feature maps \(\boldsymbol{F}_1\). Additionally, we integrate an image classification backbone, such as ResNet~\cite{resnet} or ViT~\cite{vit} that produces another set of feature maps \(\boldsymbol F_2\). Lastly, we combine the feature maps \(\boldsymbol{F}_1 \text{ and } \boldsymbol{F}_2\) in a learned fashion to forge a unified feature representation \(\boldsymbol{F} = \textsc{concat}(\boldsymbol F'_1, \boldsymbol F'_2)\).

\subsubsection{Path Prediction}
\label{subsec:path-predictor}

The path predictor is the core component of our architecture and utilizes the DETR framework~\cite{detr}. The path predictor consists of two stages. In the encoder stage, the image features $\boldsymbol F$ are encoded using an iterative self-attention mechanism. In the following decoder stage, this encoded version of the extracted features is then used to transform a predefined number $n_Q$ of lane path queries into proposal vectors using self- and cross-attention. Finally, two prediction heads predict all $n_Q$ path proposals in parallel. In the following, we detail the specific components:

\noindent\textbf{Transformer Encoder:} To maintain the positional relationships within the image features before feeding them into the transformer encoder, we apply a fixed two-dimensional sinusoidal positional encoding $\boldsymbol P_{enc}$~\cite{posencoding}.
% \begin{align}
%     \scriptsize
%     \boldsymbol P_{enc}(x, y, z) =
%     \begin{cases}
%          \sin(x/10000^{4i/c}) & \text{if } z = 2i\\
%          \cos(x/10000^{4i/c}) & \text{if } z = 2i + 1\\
%          \sin(y/10000^{4i/c}) & \text{if } z = 2i + c / 2\\
%          \cos(y/10000^{4i/c}) & \text{if } z = 2i + 1 + c/2\\
%     \end{cases} \\
%     \text{for any } i \in [0, c / 4) \nonumber
% \end{align}
We add the positional encoding onto the original features and flatten the feature maps $\boldsymbol Z =  \textsc{flatten}(\boldsymbol F + \boldsymbol P_{enc})$ and feed them to the transformer encoder to yield $\boldsymbol Z'$.

\noindent\textbf{Transformer Decoder:} The decoder takes the encoded image features $\boldsymbol Z'$ to generate a set of path proposals. We employ a set of fixed-size vectors \(\boldsymbol Q = (\boldsymbol q_1, \ldots, \boldsymbol q_{n_Q})^T\) as trainable transformer queries, where each query $\boldsymbol q \in \boldsymbol Q$ is a higher-dimensional representation of a potential lane path. This query design aims to capture various paths without specifying the order of prediction for each path, eliminating the need for positional encoding in the decoder segment of our model. By employing a standard decoder structure with $n_{dec}$ layers, we refine each path query $\boldsymbol q \in \boldsymbol Q$ into a path proposal vector $\boldsymbol q' \in \boldsymbol Q'$.

\noindent\textbf{Prediction Heads:} For each path proposal vector \(\boldsymbol q' \in \boldsymbol Q'\), we obtain the following outputs using two distinct prediction heads: First, the detection head predicts the likelihood that the proposed path exists using an MLP followed by a Sigmoid activation layer to obtain outputs in \([0, 1]\). Secondly, the control head predicts the configuration of each path as a series of control points \(\boldsymbol A = (\boldsymbol a_1, \ldots, \boldsymbol a_{n_{cp}})^T \). The MLP-based control head projects the points' coordinates onto a normalized $[0, 1]$ scale wrt.\ \(\boldsymbol I_{target}\). This process is executed concurrently for all path proposals \(\boldsymbol q' \in \boldsymbol Q'\), producing a collection of path with associated probabilities \(\hat Y = \{(l_i, \boldsymbol A_i)\}_{i = 1}^{n_Q}\) as shown in Fig.~\ref{fig:teaser}.

\subsection{Training}
\label{subsec:training}
Optimizing the ALGT model is challenging due to varying set sizes between the ground truth and predicted paths. First, we identify the optimal matching between the predicted paths $\hat Y$ and the ground truth paths $Y$ using Hungarian matching~\cite{hungarian-matching}:
\begin{equation}
    \mathcal C_{match}(Y_i, \hat Y_j) = \sum_{k = 1}^{n_{cp}} \alpha \cdot d(\boldsymbol y_i^k,  \boldsymbol a_j^k) + \beta \cdot (1 - l_j),
\end{equation}
where $d(\boldsymbol y_i^k,  \boldsymbol a_j^k)$ represents the Manhattan distance between the $k$-th control points of $Y_i$ and $\hat Y_j$, while $\alpha$ and $\beta$ are constants that balance the contributions of spatial discrepancies and classification confidence, respectively. Following the obtained optimal matching, we minimize a composite loss that penalizes the regression errors of control points and the classification inaccuracies of the predicted paths:
\begin{equation}
    \mathcal L = \alpha \cdot \sum_{Y_i \in Y} \mathcal L_{mse}(Y_i, \hat Y_{\sigma^*(i)}) + \beta \cdot \mathcal L_{bce}(l_i),
\end{equation}
where the MSE loss regularizes the errors over all control points and the binary cross-entropy penalizes path likelihoods.

\subsection{Path Filtering and Aggregation}
\label{subsec:path-aggregation}
In the aggregation step of our approach, we convert the set of path proposals \(\hat{Y}\) into the successor lane graph representation \(\hat G\). First, we disregard all path proposals that do not meet a certain minimum likelihood threshold \(p_{min}\). Next, we fuse the thresholded paths into a cohesive successor lane graph \(\hat G\) by adding directedness and merging the individual path graphs iteratively: Given a candidate path graph to merge we identify the nearest nodes \(v^*\) in the existing graph using L2 distances. If the distance \(d(v, v^*)\) is less than or equal to \(d_{max}\), indicating the nodes represent the same location, we merge \(v\) into \(v^*\). Otherwise, \(v\) is added as a new node. After completing the iterative merging process, we obtain our final successor lane graph, \(\hat{\mathcal{G}}\), as shown in Fig.~\ref{fig:teaser}.

\section{Experimental Evaluation}\label{sec:experiments}

\begin{figure*}
\renewcommand{\arraystretch}{2}
\centering
\footnotesize
\setlength{\tabcolsep}{0.1cm}% for the horiz padding
\begin{tabular}{m{1.6cm}m{15.0cm}}
     {\raggedleft Ground Truth} & \includegraphics[width=0.19\textwidth]{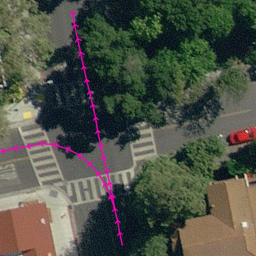}\hfill
    \includegraphics[width=0.19\textwidth]{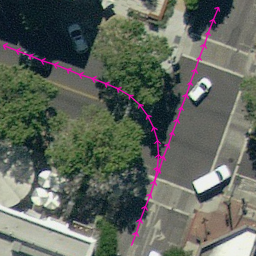}\hfill
    \includegraphics[width=0.19\textwidth]{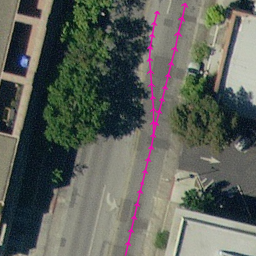}\hfill
    \includegraphics[width=0.19\textwidth]{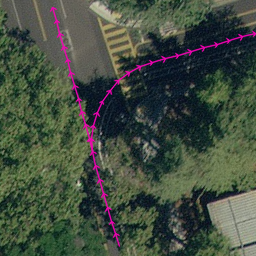}\hfill  \\
    {\raggedleft ALGT (ours)} & \includegraphics[width=0.19\textwidth]{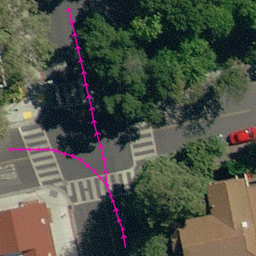}\hfill
    \includegraphics[width=0.19\textwidth]{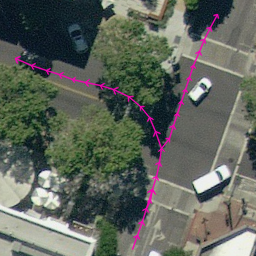}\hfill
    \includegraphics[width=0.19\textwidth]{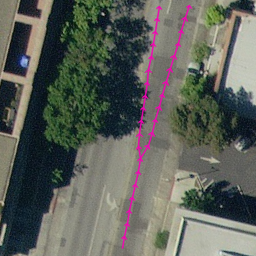}\hfill
    \includegraphics[width=0.19\textwidth]{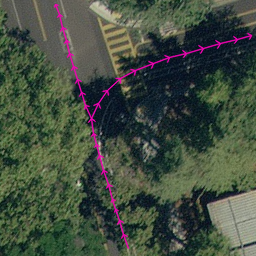}\hfill\\  \\
    \end{tabular}
    \vspace{-0.3cm}
    \caption{Qualitative results obtained by our ALGT model in comparison to the ground truth. The top row represents the ground truth, while the bottom row presents our model's predictions. Our proposed architecture predicts highly accurate lane graphs that do not suffer from sampled node positions as LaneGNN and shows high split detection accuracy. In general, the obtained graphs show a smooth characteristic.}
    \label{fig:qualitative-results}
\end{figure*}

In the following, we present our experimental findings on the Palo Alto split of the \textit{UrbanLaneGraph} benchmark dataset~\cite{lane-gnn}. To quantify our experimental results we make use of the same set of metrics utilized in the benchmark: TOPO and GEO measuring geometric and topological similarity, the average path length similarity (APLS), the split detection accuracy (SDA), and the image-based Graph IoU. We present quantitative evaluations including an ablation study and compare our proposed ALGT model with the strong LaneGNN baseline~\cite{lane-gnn}. While the LaneGNN method shows high topological accuracy it suffers from inaccurate node positions, which show significant offsets with respect to the ground truth graphs. We refer to the LaneGNN paper~\cite{lane-gnn} for more insights. Additionally, we also present qualitative results in Fig.~\ref{fig:qualitative-results}.

\begin{table}[htbp]
    \centering
    \setlength{\tabcolsep}{2.6pt}
    {\scriptsize
    \begin{tabular}{l|cccccc}
        \toprule
        Variant & TOPO P/R & GEO P/R & APLS & SDA\textsubscript{20} & SDA\textsubscript{50} & Graph IoU \\
        \midrule
        \textbf{Path Representation}&&&&&&\\
        \midrule
        Bézier&0.395/0.339&0.567/0.527&0.619&0.191&0.405&0.290\\
        \underline{Polyline}&\textbf{0.479}/\textbf{0.420}&\textbf{0.639}/\textbf{0.594}&\textbf{0.664}&\textbf{0.251}&\textbf{0.479}&\textbf{0.338}\\
        \midrule
        \textbf{Backbone}&&&&&&\\
        \midrule
        ResNet-50&0.268/0.223&0.433/0.396&0.509&0.158&0.367&0.200\\
        ViT-B-16&0.253/0.212&0.418/0.383&0.465&0.114&0.348&0.195\\
        PSPNet &0.479/\textbf{0.420}&0.639/\textbf{0.594}&0.664&\textbf{0.251}&\textbf{0.479}&0.338\\
        \underline{PSPNet + ResNet-50} &\textbf{0.485}/0.414&\textbf{0.644}/0.587&\textbf{0.665}&0.237&0.447&\textbf{0.345}\\
        \midrule
        \textbf{Architecture}&&&&&&\\
        \midrule
        $(1,1,64,10)$&0.432/0.353&0.598/0.536&0.643&0.193&0.344&0.304\\
        $(2,2,128,20)$&0.474/0.402&0.631/0.575&0.651&0.223&0.420&0.331\\
        \underline{$(4,4,128,10)$}&\textbf{0.479}/\textbf{0.420}&\textbf{0.639}/\textbf{0.594}&\textbf{0.664}&\textbf{0.251}&\textbf{0.479}&\textbf{0.338}\\
        \bottomrule
    \end{tabular}}
    \caption[Ablation studies of our model.]{Ablation studies conducted on our foundational ALGT model, utilizing the Palo Alto validation set from the UrbanLaneGraph dataset \cite{lane-gnn}. The underlined variants indicate the selected parameters used for comparison again LaneGNN. The highest-performing values in each category are highlighted in bold. The architecture ablation study compares varying configurations of the number of encoder and decoder layers, the model's hidden dimensionality and the number of queries: $(n_{enc}, n_{dec}, c, n_Q)$. Higher values are better, best results are written bold.}
    \label{tab:ablation-results}
\end{table}

\subsection{Quantitative Results}
We conduct an ablation study on the utilized path representation, different image backbones, and different encoder-decoder sizes of the network. As detailed in Tab.~\ref{tab:ablation-results}, we demonstrate that the polyline-based prediction greatly outperforms the Bézier parametrization across all metrics, which leads us to conclude its greater suitability for set-based lane graph prediction tasks. Moreover, we observe that the PSPNet image backbone outperforms variants that solely utilize standard image backbones such as ResNet or ViT that are often rather used in detection tasks. When coupling a PSPNet instance with a ResNet-50 as described in Sec.~\ref{subsec:backbone}, we observe similar results compared to only relying on a PSPNet instance.
Furthermore, a variation of the transformer encoder and decoder sizes shows that a higher number of encoder and decoder layers improves predictions while keeping the number of path proposals at 10.

As depicted in Tab.~\ref{tab:test-set-evaluation}, we observe competitive results on the Palo Alto test data split when comparing our ALGT model with the LaneGNN baseline~\cite{lane-gnn}. While LaneGNN shows greater topological accuracy, the ALGT model vastly outperforms the LaneGNN on the APLS metric. This outlines that the ALGT architecture does not suffer from inaccurate node positions and ultimately resolves this limitation inherent to sampled node manifolds used by LaneGNN.

\begin{table}[htbp]
    \centering
    \setlength{\tabcolsep}{2.6pt}
    {\footnotesize
    \begin{tabular}{l|cccccc}
        \toprule
        Method & TOPO P/R & GEO P/R  & APLS  & SDA\textsubscript{20}& SDA\textsubscript{50} & Graph IoU \\
        \midrule
        LaneGNN & \textbf{0.584}\textbf{/0.744} & 0.582/\textbf{0.739} & 0.177 & 0.220 & 0.367 & \textbf{0.378}\\
        % ALGT (ours) & 0.457/0.422 & \textbf{0.629}/0.597 & \textbf{0.718} & 0.204 & \textbf{0.471} &0.320\\ % resnet only
        ALGT & 0.481/0.437 & \textbf{0.645}/0.606& \textbf{0.714} & \textbf{0.224} &\textbf{0.497} & 0.343\\ % resnet+psp
        \bottomrule
    \end{tabular}}
    \caption[Quantitative results of our model and LaneGNN \cite{lane-gnn}.]{Quantitative results of our ALGT model in comparison with the LaneGNN \cite{lane-gnn} baseline mode for the successor lane graph prediction task. Both models are trained and tested on the subset of Palo Alto from the UrbanLaneGraph dataset \cite{lane-gnn}. Higher values are better, best values written bold.}
    \label{tab:test-set-evaluation}
    
\end{table}

\subsection{Qualitative Results}
We present additional qualitative results of the proposed ALGT model in Fig.~\ref{fig:qualitative-results}. As depicted, our proposed method predicts highly accurate lane graphs that do not suffer from sampled node positions (as LaneGNN) and shows high split detection accuracy. In addition, we show a number of failure cases in Fig.~\ref{fig:qualitative-failure} in the appendix.

\section{Conclusion}
We presented a novel successor lane graph prediction approach that generates highly accurate paths while not suffering from graph initialization errors. This leads to improved path accuracy and allows better split point predictions at intersections. Future work could address the temporal aggregation of the transformer-based predictions and tackle the out-of-distribution problem inherent to large-scale lane graph prediction.

\begin{figure*}
    \centering
    \footnotesize
    \begin{tabular}{m{1.6cm}m{15cm}}
     {\raggedleft Ground Truth} & \includegraphics[width=0.2\textwidth]{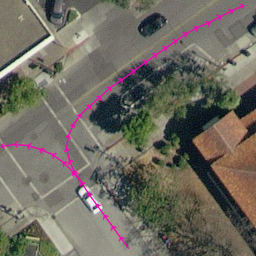}\hfill
    \includegraphics[width=0.2\textwidth]{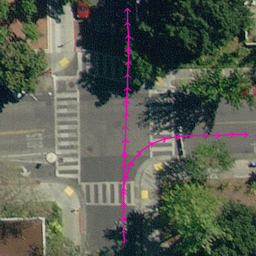}\hfill
    \includegraphics[width=0.2\textwidth]{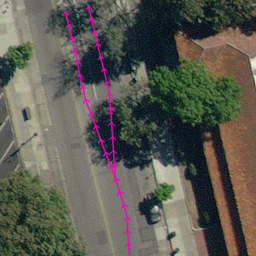}\hfill
    \includegraphics[width=0.2\textwidth]{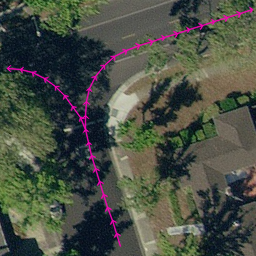}\hfill  \\
    {\raggedleft ALGT (ours)} & \includegraphics[width=0.2\textwidth]{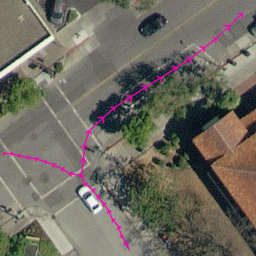}\hfill
    \includegraphics[width=0.2\textwidth]{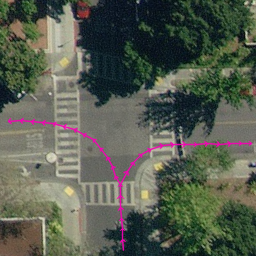}\hfill
    \includegraphics[width=0.2\textwidth]{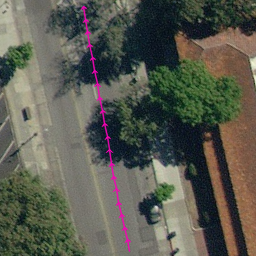}\hfill
    \includegraphics[width=0.2\textwidth]{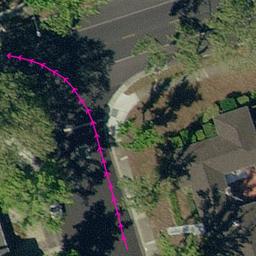}\hfill\\
    \end{tabular}
    \caption{Failure cases of the ALGT model compared to GT}
    \label{fig:qualitative-failure}
\end{figure*}

%% Use plainnat to work nicely with natbib.
\footnotesize
\bibliographystyle{plainnat}
\bibliography{references}

% \appendix
% \clearpage

\section{Appendix}
\normalsize

In addition to the qualitative insights shown in the main manuscript, we provide a number of failure cases in Fig.~\ref{fig:qualitative-failure}.

\end{document}